\documentclass{article}

\usepackage{microtype}
\usepackage{graphicx}
\usepackage{amsmath}
\usepackage{amssymb}
\usepackage{booktabs}
\usepackage{tabularx}
\usepackage[pagebackref=true,breaklinks=true,colorlinks,bookmarks=false]{hyperref}
\renewcommand*{\thesection}{\Roman{Section}}

\usepackage[accepted]{icml2021}

\icmltitlerunning{XAI Handbook: Towards a Unified Framework for Explainable AI}

\begin{document}

\twocolumn[
\icmltitle{XAI Handbook: \\
           Towards a Unified Framework for Explainable AI}



\icmlsetsymbol{equal}{*}

\begin{icmlauthorlist}
\icmlauthor{Sebastian Palacio}{equal,wrk,ed}
\icmlauthor{Adriano Lucieri}{equal,wrk,ed}
\icmlauthor{Mohsin Munir}{wrk,ed}
\icmlauthor{J\"orn Hees}{wrk}
\icmlauthor{Sheraz Ahmed}{wrk}
\icmlauthor{Andreas Dengel}{wrk,ed}
\end{icmlauthorlist}

\icmlaffiliation{wrk}{German Research Center for Artificial Intelligence (DFKI GmbH), Kaiserslautern, Germany}
\icmlaffiliation{ed}{TU Kaiserslautern, Kaiserslautern, Germany}

\icmlcorrespondingauthor{Sebastian Palacio}{sebastian.palacio@dfki.de}
\icmlcorrespondingauthor{Adriano Lucieri}{adriano.lucieri@dfki.de}
\icmlcorrespondingauthor{Mohsin Munir}{mohsin.munir@dfki.de}
\icmlcorrespondingauthor{Sheraz Ahmed}{sheraz.ahmed@dfki.de}
\icmlcorrespondingauthor{J\"orn Hees}{joern.hees@dfki.de}
\icmlcorrespondingauthor{Andreas Dengel}{andreas.dengel@dfki.de}

\icmlkeywords{Machine Learning, XAI, Explainability, Interpretability, Artificial Intelligence, Computer Vision, Healthcare}

\vskip 0.3in
]



\printAffiliationsAndNotice{\icmlEqualContribution} 

\begin{abstract}
The field of explainable AI (XAI) has quickly become a thriving and prolific community.
However, a silent, recurrent and acknowledged issue in this area is the lack of consensus regarding its terminology.
In particular, each new contribution seems to rely on its own (and often intuitive) version of terms like ``explanation'' and ``interpretation''.
Such disarray encumbers the consolidation of advances in the field towards the fulfillment of scientific and regulatory demands e.g., when comparing methods or establishing their compliance w.r.t.~biases and fairness constraints.

We propose a theoretical framework that not only provides concrete definitions for these terms, but it also outlines all steps necessary to produce explanations and interpretations.
The framework also allows for existing contributions to be re-contextualized such that their scope can be measured, thus making them comparable to other methods.

We show that this framework is compliant with desiderata on explanations, on interpretability and on evaluation metrics.
We present a use-case showing how the framework can be used to compare LIME, SHAP and MDNet, establishing their advantages and shortcomings.
Finally, we discuss relevant trends in XAI as well as recommendations for future work, all from the standpoint of our framework. 
\end{abstract}

The growing demand for explainable methods in artificial intelligence (a.k.a. \emph{eXplainable AI} or XAI) has recently caused a large influx of research on the subject~\cite{arrieta2020explainable}.
As AI systems are being steadily adopted for more and more high-stake decisions like loans~\cite{bussmann2020explainable}, access to medical care~\cite{rudin2018optimized} or security of in-vehicular networks~\cite{kang2016intrusion}, stakeholders depending on those decisions are beginning to require justifications, similar to those provided by humans.
However, despite the apparent simplicity of the problem at hand, proposed solutions have spanned into multiple niches simply because they are based on different definitions for terms like ``explanation'' or ``interpretation''.
An increasing number of contributions often rely on their own, and often intuitive notions of ``explainability'' or ``interpretability''--a phenomenon dubbed ``the inmates running the asylum''--which eventually leads to failure to provide satisfactory explanations~\cite{miller2017inmates}.
This self-reliance in the goal's definition has caused confusion in the machine learning community, as there is no agreed upon standard which can be used to judge whether a particular model can be deemed ``explainable'' or not.
A lack of consensus has resulted in research that, albeit exciting, ends up tackling different problems, and hence cannot be compared~\cite{bibal2016interpretability,lipton2018themythos}.
This issue has been accentuated by what the term ``explanation'' refers to in the context of AI.
Is it an approximation of a complex model~\cite{al2020contextual,lundberg2017unified}, an assignment of causal responsibility~\cite{miller2019explanation}, a set of human intelligible features contributing to a model's prediction~\cite{montavon2018methods} or a function mapping a less interpretable object into a more interpretable one~\cite{ciatto2020abstract}?
To make matters worse, a variety of circular definitions for ``explanation'' can be found, alluding to further concepts like ``interpretation'' and ``understanding'' which are, in turn, left undefined.

Critiques for some of these definitions have recently emerged, arguing that they are either not suitable for high-stake decisions~\cite{rudin2019stop}, or that they are too vague to be operational~\cite{doshi2018considerations} or falsifiable~\cite{leavitt2020falsifiable}.
Moreover, unless there is an agreed upon notion of what these terms refer to, there will be a disconnect between scientific contributions and fulfillment of legal requirements such as the European General Data Protection Regulation (GDPR)~\cite{eu2018reform}.
Their so called ``Right to an Explanation'' already reflects this gap on the ambiguity of its language, opening up the possibility of bypassing\footnote{In automated decision making, explanation of individual decisions can be avoided by assuring informed consent combined with additional safeguards such as human intervention and possible contestation of decisions.} the regulation entirely~\cite{schneeberger2020european}. 
Therefore, regulatory amendments planned by the EU can substantially benefit from a consensual definition of XAI.

Negative effects of lacking a consolidated language when comparing scientific contributions have recently impacted a close field of XAI: adversarial attacks.
Upon their discovery~\cite{szegedy2013intriguing} (i.e., small, additive perturbations in the image domain that are imperceptible to humans, while causing ML models to issue arbitrary predictions), a large community quickly grew around this issue, trying to find methods that could defend against said perturbations.
Without a clear definition of what was (and what was not) considered adversarial, together with a free choice of the threat model (i.e., what the attacking agent has access to when generating adversarial perturbations), many proposed solutions have been quickly proven ineffective~\cite{athalye2018obfuscated,carlini2019evaluating}.
In turn, this community is now striving for explicit definitions of the threat model, their definition of ``small'' perturbation and therefore, what counts as an adversarial attack or not.

In order to measure and compare progress in the field of XAI, we argue that a unified foundation is vital, and that such foundation starts with an adequate definition of the field's terminology.
In particular, we propose a framework based around atomic notions of ``explanation'', and ``interpretation'' in the context of AI (with a special focus on ML and applications in computer vision).
We show how further concepts that have popped up in the literature can be rephrased in relation to our proposed definition of ``explanation'' and ``interpretation'', therefore facilitating their comparison, and extent by which they apply to methods claiming to be explainable.

\section{Related Work}
\label{sec:relatedwork}
The rapid adoption of notions of ``explainability'' and ``interpretability'' in machine learning, prompted theoreticians to step back and ponder what the extent of those is, along with their differences and similarities.
Today, there is no shortage on theoretical ideas addressing e.g., desiderata for explanations and the relation they bear with further concepts like interpretation, faithfulness, trust, etc.
As mentioned before, a lack of agreement regarding terminology, makes it impossible to list and compare literature solely based on what has been called ``explanation''.
For an abridged recount of the AI literature defining (aspects of) either term, including definitions and perspectives, see \autoref{tab:xaidefs}.
A comprehensive meta-review of XAI methods, including an extensive section on fundamental theory can be consulted in~\cite{vilone2020explainable}.

In this section, we focus on underlining some of the limitations in the \emph{scope} of recurrent trends when defining these notions, allowing for acquiescent comparisons when analysing the terminology, establishing disparities and commonalities.

Early work, stemming mostly from philosophy, gravitated around the idea of ``explanation'' as a perennial carrier of causal information~\cite{lewis1986causal,josephson1996abductive}.
Although causality plays a fundamental role for explanations (and interpretations), it is not indispensable, meaning that there are non-causal questions that can (and should) be answered by explanations~\cite{lombrozo2006structure,miller2019explanation}.
We subscribe to the latter thesis in an effort to guarantee the universality of our proposed framework, while allowing for explicit causal arguments to fit within.

More recently, explanations adopted the form of additional models that approximate the feature space of the original model~\cite{lundberg2017unified,ribeiro2016should,lakkaraju2019faithful}.
Such paradigm has encountered some push-back due to the lack of faithfulness (i.e., reliance on the same feature basis)~\cite{rudin2019stop} and its susceptibility to malicious attacks.
For example, LIME~\cite{ribeiro2016should} and SHAP~\cite{lundberg2017unified} are popular linear approximation methods that have often been used in high-stake scenarios including medical applications~\cite{de2020evolved, carrieri2020explainable}.
Despite their wide-spread use, they are now known to be easily fooled by networks trained using a malicious scaffold~\cite{slack2020fooling}.
This result shows how even linear approximations may fail at providing explanations that align with the original model\footnote{Unsurprisingly, non-linear approximations have also been proven vulnerable~\cite{dombrowski2019explanations,ghorbani2019interpretation}}.

Lastly, we find literature that concentrates on the ethos of XAI with a focus on applications in ML.
Most acknowledge the epistemological leniency when talking about ``explanations'' and terms alike~\cite{montavon2018methods,lipton2018themythos,xie2020explainable}.
A recurring motif from this literature is also to define ``explanations'' or ``interpretations'' as an agglomeration of different, more specific terms like confidence, transparency and trust~\cite{xie2020explainable, dam2018explainable, doshi2018considerations}.
Instead of offering an actionable definition, some work focuses on classifying the requirements that an explainable system should meet~\cite{xie2020explainable, lipton2018themythos} or the kind of evaluations through which a model can be deemed explainable~\cite{lipton2018themythos, doshi2018considerations}.

Although contributions from these publications add invaluable insights to the field of XAI, there is still a distinct lack of cohesion between them.
They each propose their own categorization, desiderata and general guidelines for evaluation, effectively hampering the consolidation of each individual contribution into a unified theory of XAI.

\begin{table*}[th]
\centering
\caption{Definitions of ``explanation'' and ``interpretation'' found in XAI literature.}
\begin{tabular}{@{}rp{18em}p{16em}@{}}
\toprule
\multicolumn{1}{r}{\textbf{Source}} &
  \textbf{Explanation} &
  \textbf{Interpretation} \\
\midrule
\cite{lewis1986causal} &
  ``someone who is in possession of some information about the causal history of some event (\dots) tries to convey it to someone else.'' &
  - \\
\cite{josephson1996abductive} &
  ``assignment of causal responsibility'' &
  - \\
\cite{lombrozo2006structure} &
  ``central to our sense of understanding and the currency in which we exchange beliefs. Explanations often support the broader function of guiding reasoning.'' &
  - \\
\cite{biran2017explanation} &
  - &
  ``the degree to which an observer can understand the cause of a decision'' \\
\cite{lundberg2017unified} &
  ``interpretable approximation of the original [complex] model'' &
  - \\
\cite{montavon2018methods} &
  ``collection of features of the interpretable domain, that have contributed for a given example to produce a decision (e.g., classification or regression)'' &
  ``mapping of an abstract concept (e.g., a predicted class) into a domain that the human can make sense of'' \\
\cite{dam2018explainable} &
  ``measures the degree to which a human observer can understand the reasons behind a decision (e.g., a prediction) made by the model'' &
  - \\
\cite{doshi2018considerations} &
  - &
  ``to explain or to present in understandable terms to a human'' \\
\cite{lakkaraju2019faithful} &
  - &
  ``quantifies how easy it is to understand and reason about the explanation. Depends on the complexity of the explanation'' \\
\cite{vilone2020explainable} &
  ``the collection of features of an interpretable domain that contributed to produce a prediction for a given item'' &
  ``the capacity to provide or bring out the meaning of an abstract concept'' \\
\cite{schmid2020mutual} &
  ``in human–human interaction, explanations have the function to make something clear by giving a detailed description, a reason, or justification'' &
  - \\
\cite{al2020contextual} &
  ``local approximation of a complex model [by another model]'' &
  - \\
\bottomrule 
\end{tabular}
\label{tab:xaidefs}
\end{table*}

\section{Context and Definitions}
\label{sec:framework}
In order to lay down a sound and inclusive foundation, we start by looking at core aspects that most research in XAI share, but also at what makes them incompatible.
Once more, the lack of consensus on what ``explanation'' means plays an instrumental role.
In particular, each new method has to define the \emph{context} in which it is relevant, and where its outcome can thus be applied.
What counts as an atomic notion and what is treated as a system, has been the prerogative of each scientific contribution.
This has been in part acceptable, given how under-specified modern ML tasks are~\cite{lombrozo2006structure,d2020underspecification}.
Starting from an intuitive idea of object classification, we already assume and accept the interaction of signs, objects and interpreters from semiotic theory~\cite{sowa1983conceptual}.
When working on the image domain, limitations coming from the representation gap~\cite{smeulders2000content} are not taken into consideration\footnote{To some extent, indirectly addressed by the growing scale of datasets like Imagenet~\cite{russakovsky2015imagenet} or MS-COCO~\cite{lin2014microsoft}}, neither the ambiguity of annotations stemming from the semantic gap~\cite{smeulders2000content} or even the teleological assessment of modern supervised classifiers (i.e., they work because the function meets the expectations, regardless of how they do it). 
These requirements get further reduced to low level mathematical primitives e.g., the notion of a ``chicken'' gets represented as a set of points $\mathbf{x}\in \mathbb{R}^{d_\mathbf{x}}$ which get further simplified as tensors of bytes in the range $[0, 255]$.

\begin{figure}[t]
\begin{center}
\includegraphics[width=\linewidth]{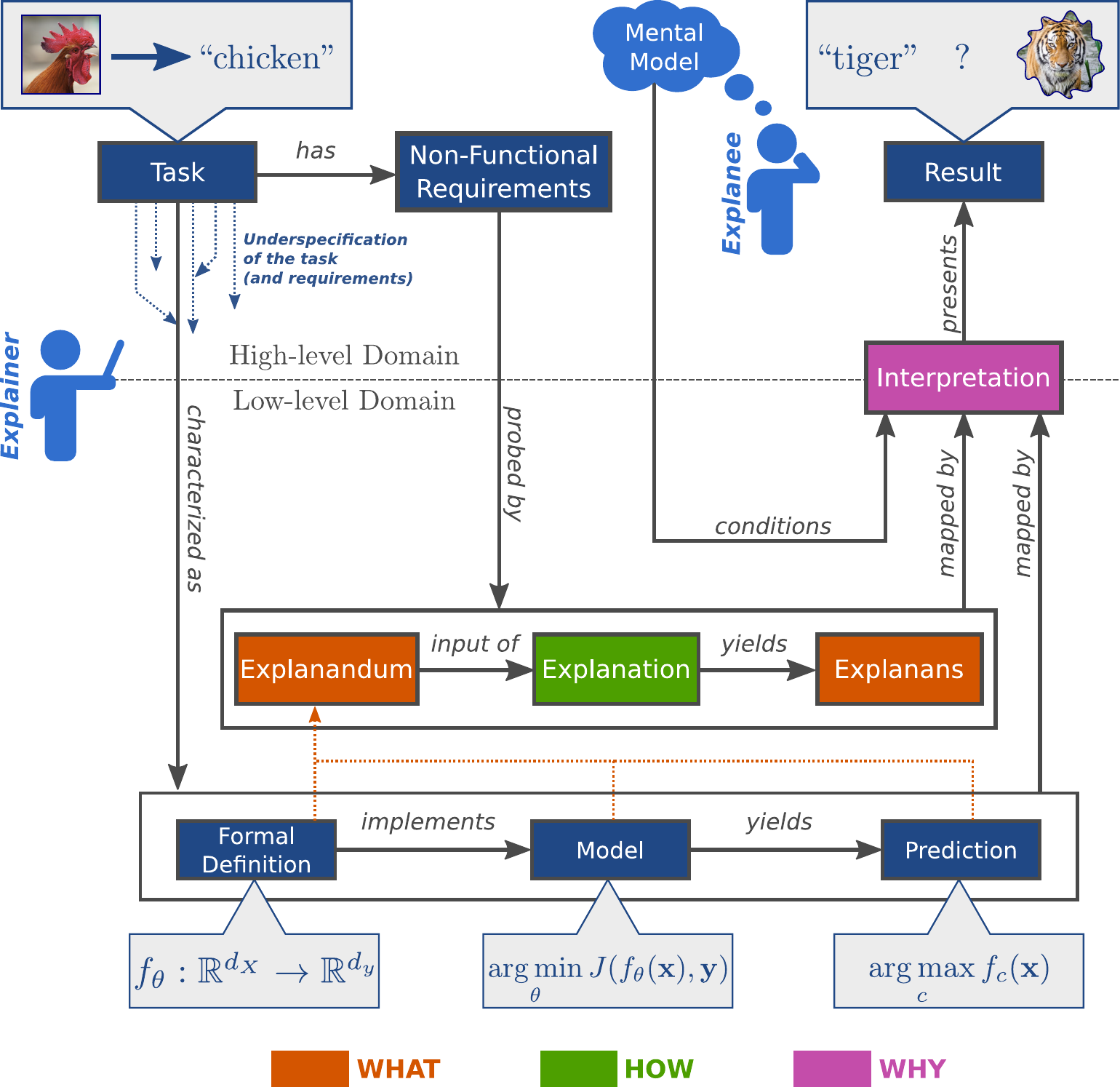}
\end{center}
\caption{Overview of our proposed framework. A task defined in the high-level domain gets an under-specified characterization, leaving out non-functional requirements. ``Explanations'' are methods that probe for said requirements. Interpretations are mappings from low- to the high-level domain.}
\label{fig:overview-large}
\end{figure}

In turn, XAI is essentially searching for evidence about non-functional requirements of the high-level task (e.g., whether a higher relevance score is being attributed to the area where a target object is) within the low-level primitives such as tensors, probabilities, and model parameters (\autoref{fig:overview-large}).
The way we accept mathematical distributions as evidence for the presence (or absence) of an object in an image follows a well-defined mapping from high-level ideas to low-level primitives.
Now, in the absence of well-established mappings between the task's non-functional properties and its corresponding low-level primitives, we are obliged to define one explicitly.

To that end, we propose a framework to establish mappings for non-functional properties, such that existing work is covered by it while providing the required rigor to serve as a vehicle for scientific discussion.
We begin by identifying two fundamental characteristics that such a framework must have:
\begin{enumerate}
    \item \textbf{Commensurable:} in order to fairly compare two different methods, common measures need to exist. A common vocabulary is therefore required, upon which these metrics can operate. In particular, what counts as an ``explanation'' and what is ``interpretation'' needs to be agreed upon beforehand.
    \item \textbf{Universal:} a generic workflow has to exist, defining the context that identifies atomic primitives, and operations on those primitives. Comparisons between primitives and operations are possible as long as the context in which they are being compared remains the same.
\end{enumerate}

We begin by looking at general notions and definitions for the terms ``explanation'' and ``interpretation'' as a basis for a refined characterization of said terms in the context of XAI.
Moreover, we will identify the minimal requirements for defining a context and therefore, for bounding the scope of explanations and interpretations.  

\subsection{Towards Commensurable Explainability: a Common Vocabulary}
\label{sec:commensurable}
Most definitions treat ``explanation'' and ``interpretation'' in a similar fashion, sometimes even as synonyms.
However, there are subtle but fundamental differences that allow some initial distinction to be drawn between them.
First, we need to ask what kind of semantic entity an explanation is: is it an action, an outcome, a process or an object?
For most textbook definitions (see \autoref{tab:dictdefs}), explanations are seen as statements.
In turn, such statements are nothing but descriptions \emph{about} an already existing entity (the \textit{explanandum} or the one which is subject to description).
From a functional perspective, an explanation is therefore the process by which an explanandum is described.
Finally, to avoid confusions between the process of explaining and its output, we refer to the former as \textit{explanation} and the latter as the \textit{explanans}.

To prevent any kind of circular definition, the explanandum needs to exist axiomatically, thus it has to refer to objects or symbols that are self-evidently true.
In other words, we require explanandums to be factual and axiomatic.

As we seek explanations for AI models, we find such suitable facts in the form of low-level mathematical primitives used to build the models themselves: support vector machines have a decision boundary equation, coordinates of the support vectors, the tolerance threshold, etc.
A Neural Network has values for each individual parameter, the equations governing how they connect with each other, the value of the cost function when computed on a particular input, the gradient that can be computed on that loss, etc.
All of these are concrete, undisputed facts (assuming there are no bugs) that are suitable explanandums.

In a simplified, more intuitive form, a first definition of ``explanation'' can be formulated as follows:

\begin{quote}
\textit{An explanation is the process of describing one or more facts.}
\end{quote}

The third aspect of an explanation deals with its purpose.
Ideally, the output of the explanation (i.e., the explanans) exposes patterns or statistics that were not evident before.
For example, overlaying the gradients of an image classifier w.r.t.~an input sample $\frac{\partial\mathcal{L}(f(\mathbf{x}), y)}{\partial \mathbf{x}}$ can locate areas that are more sensitive to changes in the input (e.g., by adding small levels of noise).
Once more, resorting to textbook definitions (\autoref{tab:dictdefs}) we see that most of them mention ``making something understandable, clear, comprehensible'' as the goal of an ``explanation''.
In other words, the purpose of an explanation is to enable (human) understanding.
Explanations are therefore bound to describe facts in a way that ultimately leads to understanding.
At this point, we can think of what consumers of explanations (i.e., explainees) need to understand: on the one hand, there are characteristics of the fact being described (e.g., location of the magnitude and sign of a gradient w.r.t.~an input sample) which may help one understand what the gradient is\footnote{Assuming that the explanee's mental model is otherwise equipped with the necessary knowledge to understand this concept.}.
On the other hand, there are some characteristics of the described fact in relation to other high-level phenomena e.g., how a high gradient value relates to a latent relevant feature.

With this in mind, we arrive at a revised definition for ``explanation'':

\begin{quote}
\textit{An explanation is the process of describing one or more facts, such that it facilitates the understanding of aspects related to said facts (by a human consumer).}
\end{quote}

The dependency between explanations and humans is explicit, as the action of understanding can be thought of as being unique to humans.
Whenever explanations are consumed by other machines (or can be otherwise executed without humans in the loop) they are no longer serving as a vehicle to explain but rather as part of a verification system.

In order to constrain the many ways a description can be interpreted, a contract must be first introduced detailing the valid meanings that can be extracted out of that description.
In other words, there should be an agreement on how to read the symbols of a description e.g., which colors on a heatmap mean high or low values. 
Such agreement anchors or \emph{assigns} meaning to a primitive entity (in our case, an explanans).

Referring back to textbook definitions for ``interpretation'', we see how most of them rely on the term ``explanation'', and thus leading to an inexorable circular definition.
The one remaining exception follows the philosophical origin of the word and already lays out the requirements of a contract by defining an ``assignment of meaning''.
In fact, the conveyance of meaning is common to all entries, in one way or another. 
We go along these lines to sketch the definition of interpretation in XAI as follows:

\begin{quote}
\textit{Interpretation is the assignment of meaning (to an explanation).}
\end{quote}

For ML, the assigned meaning refers to notions of the high-level task for which the explanans is provided as evidence.
An interpretation is therefore bridging the gap between underspecified non-functional requirements of the original task and its representation in formal, low-level primitives (e.g., high Shapley values for pixels on a chicken's beak indicate its correct detection and relation to the class ``chicken'').

In order for an explanation to fulfill its goal (facilitate understanding), the terminology and symbols known to the explainee i.e., the consumer of the explanation, have to match those used by the explainer i.e., the proponent of the explanation, when assigning meaning to an explanation.
In other words, the complexity (otherwise known as parsimony) of the interpretation should not be greater than the explainee's capabilities to fathom its meaning~\cite{ras2018explanation,sokol2020explainability}.

Sometimes, there are no fundamental reasons to prefer one interpretation over another (e.g., make red represent high values instead of blue or white) as long as one is agreed upon.
There are scenarios in which interactivity allow for correction and negotiation of the meaning being assigned to explanations (e.g., arguing for the importance of using another color other than red for high values of a heatmap).
Often, however, it is fundamental to consider the mental model of the explainee, as it expedites the process of understanding (i.e., it has an optimal degree of parsimony).
This involves estimating the relevant mental constructs known to potential explainees beforehand, as exact mental models may not be available.
In addition, mental constructs unknown to the explainee, but crucial for the interpretation, need to be properly introduced.
In the worst case, ambiguities can lead to a misinterpretation of explanations\footnote{Prominent examples have even made it into mainstream media~\cite{bbc-fbnews}.}.
A simple strategy to bridge the gap between the explainee's mental model and the constructs of an interpretation is to rely on common conventions such as those coming from a natural universal language~\cite{king2005human}.

Note that the explainee's mental model is assumed to be based on true facts; the consequences of engaging in the process of explanation and interpretation based on false premises can be catastrophic~\cite{lombrozo2006structure}.
In consequence, matching an explainee's mental model is desirable but not indispensable.
Interpretations are valid as long as their statements cohere, and the inclusion of a particular explainee's mental model should only accelerate the process of understanding, while leaving the essence of the interpretation itself unaltered.

Even if the methods to explain (or the agreements to interpret) vary, the process of understanding will always be supported by the same mechanism: description of facts followed by the assignment of meaning for the description itself.
While the meaning being assigned can be contested (e.g., as part of the scientific process where one seeks to falsify a statement) the process of assignment cannot.

\begin{table*}[th]
\centering
\caption{Definitions of ``explanation'' and ``interpretation'' according to various dictionaries. Explanation is referred to as an object while interpretation is more commonly associated to an action. Accessed on 05.01.2021.}
\begin{tabular}{@{}rp{20em}p{19em}@{}}
\toprule
\textbf{Source}           & \textbf{Explanation}           & \textbf{Interpretation}             \\
\midrule
\textit{Merriam-Webster}  & \textbf{act} of making plain or understandable
                          & \textbf{action to explain} or tell the \underline{meaning} of                                                              \\
\textit{Cambridge}        & \textbf{the details} or other information that someone gives to make something clear or easy to understand
                          & \textbf{an explanation} or opinion of what something \underline{means}                                                     \\
\textit{Oxford}           & \textbf{a statement} or account that makes something clear
                          & \textbf{the action of explaining} the \underline{meaning} of something                                                     \\
\textit{Dictionary.com}   & \textbf{statement} made to clarify something and make it understandable
                          & \textbf{explain}; action to give or provide the \underline{meaning} of; explicate; elucidate                               \\
\textit{Princeton}        & \textbf{a statement} that makes something comprehensible by describing the relevant structure or operation or circumstances etc
                          & \textbf{an explanation} of something that is not immediately obvious; a mental representation of the \underline{meaning} or significance of something  \\
\textit{Wikipedia}        & \textbf{a set of statements} usually constructed to describe a set of facts that clarifies the causes, context, and consequences of those facts
                          & A philosophical interpretation is \textbf{the assignment of \underline{meanings}} to various concepts, symbols, or objects under consideration. \\
\bottomrule
\end{tabular}
\label{tab:dictdefs}
\end{table*}

\subsection{Universal Context for XAI Methods: the \emph{what}, the \emph{how} and the \emph{why}}
\label{sec:universality}
One of the main aims of equipping ML models with explanation capabilities is to contest previous beliefs w.r.t.~a particular prediction by constraining an otherwise undetermined problem~\cite{lombrozo2006structure}.
So far, we have already constrained two core definitions in XAI, in an effort to define a unified language that allows us to engage in scientific discussion.
Said terminology still needs to fit into a more generic, procedural framework where novel and existing contributions can be placed and thus, compared.

Explanations and interpretations are ultimately aimed at answering questions arising from the underlying process (i.e., the ML model) that issued a prediction.
These questions, in their more generic form, correspond to variants of \textit{what}, \textit{how} or \textit{why} queries.
Hence, our interest lies in defining a framework that addresses these questions when defining novel explanations and interpretations.
In fact, we show that both terms are central for answering all three questions.
We describe the scope of the aforementioned questions, and the role that explanations and interpretations both play when answering them.

\textbf{The \textit{what} defines the domains in which explanations operate.}
As defined in \autoref{sec:commensurable}, we find that explanations, as processes, take in elements of a particular \textit{source} domain, and output descriptions that exist in a \textit{target} domain.
This notion of ``translation'' between two domains has been recently promoted~\cite{esser2020disentangling} although the terminology differs with the one proposed in this work.
In mathematics, whenever a function is defined, it is first expressed in terms of the domain and co-domain where the function projects values into e.g., $f: \mathbb{R}^{d_x} \rightarrow \mathbb{R}^{d_y}$.
Similarly, an explanation method should explicitly state \textit{what} is being used as input and \textit{what} is being produced as output.
Ideally, the output of the explanation (explanans) should be defined in terms of low-level primitives that are as factually true as the input (e.g., intermediate features, support vectors, gradients).  
Note that, while inputs in the \textit{source} domain are limited by the model being explained, the \textit{target} domain depends on the explanation method, where options are virtually unlimited\footnote{Practical constraints arise from the cognitive limitations of human minds (e.g., the inability to imagine a tesseract).}.

\textbf{Defining \textit{how} something is being done can be addressed from two levels of abstraction.}
First, at the system-level, there is the question of \textit{how} the model arrives at a prediction.
Second, there is the issue of \textit{how} the explanans is produced.
In other words, what are the details behind the process that transforms between domains defined by the \textit{what}, and used by the explanation.
The former is precisely the kind of inquiry that XAI methods try to answer and therefore, they cannot be included in our framework.
Instead, this kind of \textit{how} questions will be answered through methods that rely on it\footnote{In fact, all three questions--\textit{what, how, why}--have been proposed as the basis for identifying questions that are answerable through explanations~\cite{miller2019explanation}.}.
The latter on the other hand, deals with the context of explanations and their answer is essentially reduced to the explanation method itself.
Simply put, the answer of \textit{how} the explanandum is mapped to an explanans is defined by the explanation method.

As most XAI literature is devoted to the development of explanation methods (e.g., computation of relevance values~\cite{bach2015pixel, vaswani2017attention}, mapping of high-level conceptual constructs in intermediate activations~\cite{bau2017network, kim2018interpretability}), the \textit{how} is almost always thoroughly defined.
This has already nurtured insightful debates on whether linear approximations of the original model~\cite{lundberg2017unified} or even other black-boxes~\cite{al2020contextual} can be considered appropriate explanation methods~\cite{rudin2019stop,slack2020fooling}.

Together, the \textit{what} and the \textit{how} already constrain the scope in which explanations are valid.
The one remaining aspect is the context in which an explanation is interpreted.
An explanans is, by itself, only relevant within the target domain defined by the explanation.
It is the interpretation (of the explanans) which will map the low-level representations into a high-level domain where expectations regarding non-functional requirements can be validated or contested.
Consider the following interpretation: ``the result of an \textit{argmax} operation on the logits of a neural network corresponds to the predicted class''.
The high-level action of predicting a class has been mapped to an \textit{argmax} operation over a vector.
Based on this interpretation of the \textit{argmax} operation, users of neural networks can either accept this notion to gather results, or find shortcomings and propose alternative ways of fulfilling the requirement of a model to predict classes (e.g., through the refinement of the prediction through a hierarchical exclusion graph~\cite{deng2014large}).
The same principle can be applied to explanans and non-functional requirements.

Asking \textit{why} something happens, inescapably relates back to causal effects.
While these kind of relationships are among the most useful to discover (as it allows for more control over the effect by adjusting the cause), other non-causal relationships remain valuable in the toolset of explainability.
Finding out that a model is unfair, without knowing what the cause of it is, can already be helpful in high-stake scenarios (e.g., by preventing its use).
\textbf{We say that the \textit{why} relates to the nature of an interpretation.}
In short, if the interpretation bares a causal meaning, then the \textit{why} is being defined by the causal link.
If the meaning is limited to a correlation, the \textit{why} is left out of the scope of that particular interpretation.
Note that, if the explanation method is already based on causal theory~\cite{lopez2017discovering,chang2019explaining}, the assigned meaning (i.e., the link between the explanans and the high-level (non-)functional requirement) will be more direct and therefore, more likely to withstand scientific scrutiny.  
The \textit{why} is therefore not mandatory in explanations generated by XAI methods.
In any case, the explanation's context can and should be defined, be it causal or based solely on correlations.
Proponents of XAI methods are responsible for clearly stating the context in which their explanations can be interpreted.

\section{On the Completeness of the XAI Framework}
We show that our proposed framework complies with concepts and desiderata related to explanations and explainable models.
In particular, those that have been defined by Alvarez-Melis and Jaakkola~\cite{alvarez2018towards}, and Miller~\cite{miller2019explanation}.
Furthermore, we discuss evaluation metrics for XAI methods (as defined by Doshi-Velez and Kim~\cite{doshi2018considerations}) and how they also fit within our framework.

For Alvarez-Melis and Jaakkola~\cite{alvarez2018towards}, there are three characteristics that explanations need to meet: fidelity, diversity and grounding.
Fidelity alludes to the preservation of relevant information; diversity states that only a small number of related and non-overlapping concepts should be used by the explanation, while grounding calls for said concepts to be human-understandable.
In our proposed framework, fidelity is guaranteed as explanations are defined as mechanisms to describe or map inputs that are axiomatically valid using a well-defined function.
Although defining an absolute number of concepts can be rather subjective, diversity is possible by allowing a designer of explanations to purposely focus on a few concepts on the input or output of the explanation.
Grounding is essentially addressed by the interpretation, as it is the mapping from a low-level primitive to a high-level, human-understandable realm of a non-functional requirement.
Moreover, explanations require the production of artifacts specifically targeted at enabling human understanding.

Miller~\cite{miller2019explanation} has highlighted several aspects of explanations that the XAI community has been mostly unaware of: the social, contrastive and selective nature of explanations, and the irrelevance of probabilities when providing an explanation.
When defining the domain and co-domain of an explanation, there are several ways by which contrastive explanations can be offered: either several explanandums can be processed, and their explanans compared, or the explanation method itself expects multiple input pairs (possibly producing output pairs too).
Employing causal methods as explanations will inevitably encode counterfactual information (e.g., the result of applying a \textit{do}-operator) enabling a comparison with respect to the observed data.
Selective explanations closely relate to the concept of diversity from Alvarez-Melis and Jaakkola~\cite{alvarez2018towards}.
The irrelevance of probabilities mainly states that the best explanation for the average case may not be the best explanation for a particular explainee.
While true in some cases, the usefulness of tailored explanations is contingent on the use-case (as providing different explanations for two identical cases may violate fairness constraints).
Nonetheless, our framework does not preclude an explanation from doing so if the use-case calls for it.
Finally, the social aspect of explanations is reflected by its very definition, as the purpose of explanations is to ``enable \textit{human} understanding''.
Furthermore, interpretations are mappings from low-level to high-level requirements, precisely to make an explanans consumable by humans.

General guidelines for measuring explanations (or better said, their outputs) have been proposed in~\cite{doshi2018considerations}.
These are based on three kinds of evaluations depending on the scope of the application (from generic to specific) and they revolve around the involvement of automatic proxy-tasks, non-expert humans, or domain experts.
The use of automated tasks would be amenable primarily to explanations, as they already live in the realm of data structures and mathematical primitives.
Evaluations that involve humans will thus be better suited for the interpretations, as mappings to a high-level domain, where the non-functional requirements originate, ultimately affect human understanding (therefore impacting trust, confidence, etc).

A fitting example of quantifiable criteria for explanations can be found in the aforementioned work by Alvarez-Melis and Jaakkola\cite{alvarez2018towards}.
They propose three, arguably generic characteristics that their proposed explanations should meet, namely explicitness, faithfulness and stability.
We see that such properties also refer to aspects defined in our framework: explicitness or ``how understandable are the explanations'' establishes how clear the interpretation of their provided explanans (in their case, a selection of prototypes describing an expected latent feature) is.
Faithfulness or the ``true relevance of selected features'' is directly addressing the quality of an explanans via counterfactual analysis (i.e., had the selected feature not been there, would the prediction suffer any change?).
Finally, stability measures consistency of the explanation for similar inputs.
As part of the particular classification problem they work on, said property deals with an expected local Lipschitz continuity which guarantees that similar input samples will yield a similar explanans.

We see how our proposed framework offers a comprehensive language that not only aligns and encompasses previously defined desiderata regarding XAI, but also allows the identification of common ground between a wide array of concepts related to the field.

\section{Understanding the State of XAI under our Proposed Framework}
\begin{table*}[t]
\centering
\caption{Recontextualization of three popular XAI methods with the help of our proposed framework.}
\begin{tabular}{llp{12em}p{13em}p{11em}}
\toprule
\multicolumn{2}{c}{\textbf{Method}} & LIME  & SHAP & MDNet \\
\midrule
\textbf{What} & Source           & Model input \newline Model prediction & Model input \newline Model prediction & Model activations \\
\cline{2-5}
              & Target           & Linear classifier weights             & Linear classifier weights             & Word-wise attention matrix \\ 
\cline{2-5}
\textbf{How}  &                  & 1) Input perturbation sampling \newline 2) Inference on class of interest \newline 3) Training of proxy model & 1) Input perturbation sampling. \newline 2) Inference on class of interest. \newline 3) Training of unique SHAP solution for proxy model & 1) Implicit training of attention module. \newline 2) Generation of attention matrices from activations and LSTM state. \\
\cline{2-5}
\textbf{Why}  & Causal           & - & - & - \\
\cline{2-5}
              & Non-causal    & Surrogate model weights indicate the local influence of features sampled from a marginal distribution & Approximation of the average contribution of a feature to the prediction. & Approximation of the model's attention during word generation \\
\bottomrule
\end{tabular}
\label{tab:recontextualization}
\end{table*}

The proposed explanation framework helps establishing unambiguous and commensurable relations across all kinds of XAI contributions.
We demonstrate the broad applicability of our framework by re-contextualizing three popular methods, LIME, SHAP and MDNet, showing how and where they compare, while also revealing some of the gaps and complementary properties between them.

Additive feature attribution methods like LIME~\cite{ribeiro2016should} and SHAP~\cite{lundberg2017unified} are some of the most frequently used XAI techniques today.
Both were introduced as model-agnostic methods, aiming at approximating the behavior of complex models, all without requiring access to their internal variables.
On the other hand, MDNet~\cite{zhang2017mdnet} was proposed as an ``interpretable medical image network'' for diagnosis of bladder cancer through the generation of textual diagnosis along with word-wise attention maps.
At first glance, establishing a degree of commensurability (as defined in \autoref{sec:commensurable}) is not straightforward, especially when dealing with methods that operate on vastly different domains or whose explanations tackle different sets of non-functional requirements such as MDNet's and LIME's.
However, by defining the different elements of all three methods in terms of our proposed framework, it becomes possible to establish comparisons and draw limitations with respect to one another.
We discuss said elements in the remaining of this section.
For a summary of the ensuing discussion, please refer to \autoref{tab:recontextualization}.

\subsection{Source Domain:}
The explanandum of both LIME and SHAP comprises the prediction of the target model, as well as an input sample\footnote{Most \textit{local} explanation methods rely on individual samples from the model's input space as input for the explanation too. Non-local methods like TCAV~\cite{kim2018interpretability} or S2SNet~\cite{palaciofolz2018s2snets} rely on a group of samples or even a representative sample of the input space.}. 
The target model is treated as a black-box, and typically deals with low-dimensional input data.
Meanwhile, MDNet's textual and visual explanations are derived from an input sample, and from high-dimensional latent variables found within the target model.
LIME and SHAP represent the target model's low-level internal processes mainly through counterfactual analysis, while methods exposed to the model's internals can examine the flow, translation and attribution of information as it traverses the model, serving as a much more direct evidence of a model's decision process.
In fact, it has been shown that limiting access to the target model's internal representations makes the creation of adversarial attacks possible, compromising the reliability of the explanation~\cite{slack2020fooling}.
In other words, it is not enough to rely only on the input domain (of the target model) as the explanandum.

\subsection{Target Domain:}
The outputs of both LIME and SHAP consist of the weight values from linear surrogate models that are mapped to their corresponding input region (visualized as heatmaps).
Instead, MDNet's explanans is composed of a sequence of one-hot encoded words from the language model, where each word is paired with a $14\times14$ weight matrix that is spatially mapped to the input image (also visualized as a heatmap).
The combination of generated text supported by word-wise attribution maps provides a traceable structure from the model's input to the weight matrix.
While LIME and SHAP provide explanans that share some of the output space with MDNet's, the relation they bare with any internal representation of the target model is not as traceable.

\subsection{How:}
LIME and SHAP both follow similar explanation strategies that are characterized by continuously perturbing and evaluating an input sample via the target model; results are subsequently approximated through a surrogate linear model.
SHAP poses additional constrains to the surrogate's optimization and slightly differs in its sampling scheme. 
MDNet can be described as a two stage process: an image feature extraction followed by a concurrent generation of a sensitivity map (originally called ``explanation'') and textual diagnosis. 
Both the explanation and classification components of MDNet are trained simultaneously in an end-to-end fashion.
The language model for text generation is trained using diagnosis texts as the supervisory signal, fulfilling one of its functional requirements.
The word-wise attention module learns a set of sensitivity weights with additional constrains on diagnostic labels serving as indirect supervision.

\subsection{Why:}
Given all structural and low-level components of the explanations under scrutiny, as described by the \textit{what} and the \textit{how}, we now turn to the interpretation of such primitives.
The first observation is that none of the aforementioned methods provide interpretations of causal nature.
LIME and SHAP cannot provide interpretations grounded in causality due to issues related to off-manifold sampling~\cite{frye2020shapley} and the non-excludable inaccuracy when relying on proxy models~\cite{rudin2019stop}.
Nevertheless, there are non-causal interpretations worth examining.

The weight values derived from LIME's proxy models are interpreted as ``influence values''.
These values convey the relevance of individual input features w.r.t.~the prediction of the target model.
A small caveat to this interpretation is its limited validity, which applies only to the local neighbourhood around the original input sample.
SHAP's explanans, despite baring evident similarities with LIME's, allows the generation of additional global explanans through an accumulation of statistics from multiple local explanations.
Furthermore, the notion of feature attribution (i.e., how much the value of each input feature has influenced the model's prediction) inherits the properties of Shapely values.
In this case, values are interpreted as payouts to individual features, reflecting how much they contributed to the model's prediction, relative to the remaining input features.
MDNet's visual explanations are interpreted as the model's attention w.r.t.~a region within the input image while the textual-diagnosis generates a word.
In this case, we see how MDNet defines explanans as part of the model's output and not as a separate process, tying predictions and explanans together.
The correspondence between feature attribution on the image domain and the sensitivity w.r.t.~generated text comes from additional annotations available for the text.
The assumption is that latent representations of MDNet align features of image regions and text in a way that is meaningful to humans.
The link between the intuitive notion of ``attention'' and its implementation in one of MDNet's modules, has been recently contested~\cite{jain2019attention, wiegreffe2019attention}, undermining any causal relation drawn from it.

\section{Conclusions}
To address the growing heterogeneity and lack of agreement on what constitutes an explainable or interpretable model, we introduced a novel theoretical framework to consolidate research and methods developed in the field of explainable AI (XAI).
This framework is supported by two fundamental definitions, namely ``explanation'' and ``interpretation''.
These definitions are further contextualized within a general pipeline that constrains other important primitives like input/output domains, and establishes a divide between low-level mathematical constructs and the high-level, human-understandable realm of \mbox{(non-)functional} requirements.

We show that the proposed framework is compliant with desiderata regarding explanations as defined in previous work.
Moreover, existing metrics for XAI methods can be placed within the framework allowing an apples-to-apples comparison between different explanations.

Finally, we show a concrete scenario where our framework can help comparing existing XAI methods, showing the extent by which each one addresses different aspects of the explainability pipeline.

\subsection{Current Trends and Practices} In the process of defining the proposed framework, we conducted an extensive review of XAI literature.
This allowed us to identify current practices in the XAI community, some of which we summarize here in terms of our framework.

\begin{itemize}
  \item Most contributions focus on the development and use of explanation methods while neglecting the role of the interpretation (with notable exceptions like ~\cite{rudin2019stop,montavon2018methods}).
  \item Explanations are often defined without explicit definitions of the domain and co-domain (i.e., realm of the explanans and explanandum).
  \item More recently, explainable models are trying to include richer objectives through auxiliary tasks: this strategy addresses both the underspecification of the task (conveys some of the sought after non-functional requirements) and expresses some of the \textbf{invariances} that are expected of the task.
  \item In domains where AI takes over or assists human-professional workers (e.g., for medical applications), interest often shifts from the prediction to the explanation such that human experts learn and gain new insights about the task.
\end{itemize}

\subsection{Recommendations} Finally, we identify three aspects that future research in XAI should focus on in order to expedite the advancement of the current state-of-the-art.  
\begin{itemize}
  \item Before working on the specifics of an explanation, explicitly define the non-functional requirements that a task should fulfill, and that the explanation itself will be probing for. This can be achieved by defining better hypotheses e.g., one that can be falsified~\cite{leavitt2020falsifiable}.
  \item Proponents of XAI methods should be careful when addressing the complete scope of our proposed framework. In particular, a clear interpretation should be provided.
  \item Metrics for XAI methods should operate within the same level of abstraction w.r.t.~the framework i.e., compare explanans to explanans, explanandum to explanandum, interpretation to interpretation, etc.
\end{itemize}

\section*{Acknowledgements}
This work was supported by the BMBF projects ExplAINN (01IS19074), XAINES and the NVIDIA AI Lab program.

\bibliography{xai-handoook}
\bibliographystyle{icml2021}

\end{document}